\RequirePackage[svgnames]{xcolor}

\documentclass[11pt,a4paper,sharedaffil]{mystyle}

\usepackage[all]{hypcap}
\usepackage[svgnames]{xcolor}
\usepackage[comma,authoryear,compress]{natbib}
\renewcommand{\cite}{\citep}

\usepackage{algorithm}
\usepackage{algorithmicx}
\usepackage{algpseudocode}
\usepackage{microtype}
\usepackage{booktabs}
\usepackage{float}
\usepackage{bigstrut}
\usepackage{multirow}

\usepackage{amsmath}
\usepackage{amssymb}
\usepackage{mathtools}
\usepackage{amsthm}
\usepackage{mathrsfs}
\usepackage{dsfont}
\usepackage{enumitem}
\usepackage{graphicx}
\usepackage{cleveref}
\usepackage{bxcoloremoji}

\usepackage{float}
\usepackage{wrapfig}

\usepackage[utf8]{inputenc} %
\usepackage[T1]{fontenc}    %
\usepackage{hyperref}       %
\usepackage{url}            %
\usepackage{booktabs}       %
\usepackage{amsfonts}       %
\usepackage{nicefrac}       %
\usepackage{microtype}      %
\usepackage{amssymb}
\usepackage{fdsymbol}
\usepackage{wrapfig}
\usepackage{lipsum}
\usepackage{enumitem}
\usepackage{stackengine}
\usepackage[font=small,labelfont=bf]{caption}
\usepackage{color}
\usepackage{adjustbox}
\usepackage{soul}
\usepackage{rotating}
\usepackage{makecell}

\usepackage{xspace}
\usepackage{dsfont}
\usepackage{pifont}

\graphicspath{{../pics/}{pics/}}

\setlist[itemize]{nosep,leftmargin=1em,labelwidth=*,align=left}

\definecolor{codegray}{gray}{0.93}

\newcommand{\code}[1]{%
  \begingroup%
  \setlength{\fboxsep}{1.5pt}
  \colorbox{codegray}{\texttt{#1}}%
  \endgroup%
}

\definecolor{declinegreen}{RGB}{50, 160, 50} 

\newcommand{\res}[2]{#1\rlap{\scriptsize$_{\pm #2}$}}

\newcommand{\grayvar}[1]{\hspace{1em}\textcolor{black}{\textit{#1}}}

\newcommand{\Tref}[1]{Table~\ref{#1}}

\newcommand{\eref}[1]{Eq.~\eqref{#1}}

\newcommand{\Fref}[1]{Figure~\ref{#1}}

\newcommand{\Apdxref}[1]{Appendix~\ref{#1}}

\makeatletter
\DeclareRobustCommand\onedot{\futurelet\@let@token\@onedot}
\def\@onedot{\ifx\@let@token.\else.\null\fi\xspace}
\def\eg{\emph{e.g}\onedot}

\def\ie{\emph{i.e}\onedot}

\makeatother

\usepackage{soul}
\definecolor{myblue}{RGB}{218,227,243}
\definecolor{mypink}{RGB}{251,229,214}
\definecolor{myred}{RGB}{252,150,148}
\definecolor{mygreen}{RGB}{226,240,217}

\title{\textit{Beyond the Final Actor:} Modeling the Dual Roles of Creator and Editor for Fine-Grained LLM-Generated Text Detection}

\runningtitle{\textit{Beyond the Final Actor:} Modeling the Dual Roles of Creator and Editor for Fine-Grained LLM-Generated Text Detection}

\author{\href{https://www.yang-li.cn/}{\textcolor{black}{Yang Li}}}
\author{\href{https://sheng-qiang.github.io/}{\textcolor{black}{Qiang Sheng}}}
\author{\href{https://zhengjiawa.github.io/}{\textcolor{black}{Zhengjia Wang}}}
\author{\href{http://undground.fun/}{\textcolor{black}{Yehan Yang}}}
\author{\href{https://scholar.google.com/citations?user=hGZwK0cAAAAJ&hl=en}{\textcolor{black}{Danding Wang}}}
\author{\href{https://scholar.google.com/citations?user=fSBdNg0AAAAJ&hl=zh-CN&oi=ao}{\textcolor{black}{Juan Cao}}}

\affil{Institute of Computing Technology, Chinese Academy of Sciences}
\affil{University of Chinese Academy of Sciences}

\correspondingauthor{Qiang Sheng, Juan Cao}
\emails{{\{\href{mailto:liyang23s@ict.ac.cn}{liyang23s},\href{mailto:shengqiang18z@ict.ac.cn}{shengqiang18z}\}@ict.ac.cn}}

\begin{document}

\begin{abstract}
The misuse of large language models (LLMs) requires precise detection of synthetic text.
Existing works mainly follow binary or ternary classification settings, which can only distinguish pure human/LLM text or collaborative text at best. This remains insufficient for the nuanced regulation, as the LLM-polished human text and humanized LLM text often trigger different policy consequences.
In this paper, we explore fine-grained LLM-generated text detection under a rigorous four-class setting.
To handle such complexities, we propose \textbf{RACE} (Rhetorical Analysis for Creator-Editor Modeling), a fine-grained detection method that characterizes the distinct signatures of creator and editor.
Specifically, RACE utilizes Rhetorical Structure Theory (RST) to construct a logic graph for the creator's foundation while extracting Elementary Discourse Unit (EDU)-level features for the editor's style.
Experiments show that RACE outperforms 12 baselines in identifying fine-grained types with low false alarms, offering a policy-aligned solution for LLM regulation.

\vspace{3mm}

\coloremojicode{1F4C5} \textbf{Date}: April 7th, 2026

\coloremojicode{1F3E0} \textbf{Project}: \href{https://race.yang-li.cn}{https://race.yang-li.cn}

\coloremojicode{1F4AC} \textbf{Venue}: ACL 2026

\end{abstract}

\maketitle
\vspace{3mm}

\section{Introduction}
\label{sec:intro}

\begin{wrapfigure}{r}{0.42\linewidth}
  \vspace{-3em}
  \centering
  \includegraphics[width=\linewidth]{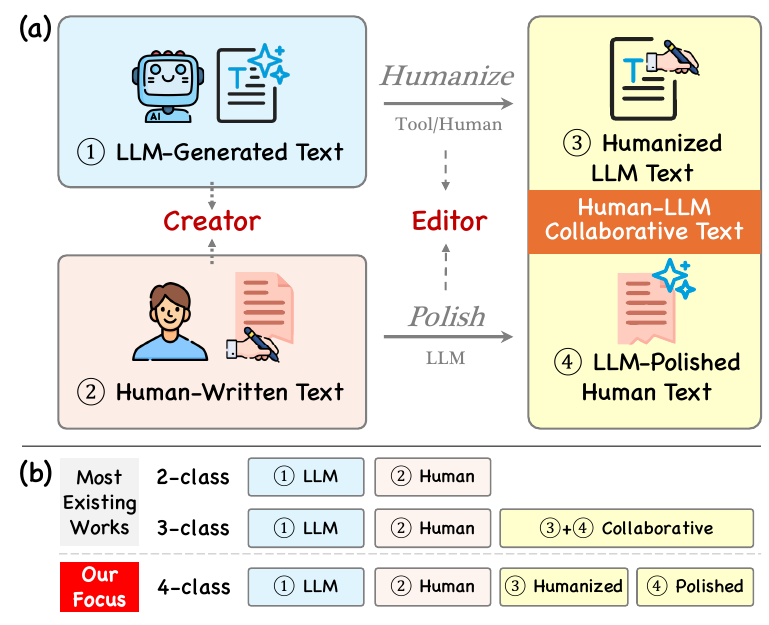}
  \caption{Illustration of our research scope. (a) A Creator-Editor framework for categorizing different types of texts in fine-grained LLM-generated text detection. (b) Comparison of the existing settings and the complex 4-class setting that we focus on in this paper.}
  \label{fig:research-scope}
  \vspace{-2em}
\end{wrapfigure}

While the surge of Large Language Models (LLMs)~\citep{openai-gpt5,yang2025qwen3} has revolutionized content creation and inspired a diverse range of downstream applications, the improper and malicious use of LLMs is also eroding the foundation of information credibility~\cite{anwar2024foundational,liu2024preventing}.
From the large-scale synthesis of misinformation~\cite{hu2025llm} to unauthorized academic assistance~\cite{ai-cheating} and LLM-based identity fraud~\cite{ai-fraud}, the ease of generating high-quality synthetic text poses a severe challenge to our trust system, necessitating effective techniques to distinguish LLM-generated text from human-written text~\citep {wu-etal-2025-survey,liu2025enhancing}.

LLM-generated text detection was primarily formulated as a binary classification task that judges whether the given text is generated by any LLM or by human~\citep{he2024mgtbench,wang-etal-2024-m4}.
However, the binary setting oversimplifies real-world scenarios where text is often a product of human-LLM collaboration~\citep{wang-etal-2025-real}.
For instance, people may ask LLMs to \textit{polish} their original drafts for better readability~\citep{yang2024chatgpt}; or conversely, \textit{humanize} LLM-generated outputs to evade detection~\citep{damage}.
Such collaborative processes yield hybrid texts that blend the characteristics of human and LLM generations, ultimately blurring the decision boundaries of conventional binary classifiers.

To address these complexities, recent studies shift towards fine-grained detection settings, typically by introducing a third ``mixed'' category~\citep{zhang-etal-2024-mixset,artemova-etal-2025-beemo,saha2025almost}. Yet, even this ternary classification remains insufficient for nuanced LLM-use regulatory policies in specific domains like academic writing~\cite{acl-policy}. Under such policies, text polishing is often considered legitimate writing assistance that requires no compulsory disclosure, while text humanizing for bypassing detectors is often prohibited, as it brings improper advantages to cheating students and damages academic integrity. 

In this paper, we study \textbf{a practically important four-class detection setting where the mixed category is explicitly separated into \textit{LLM-Polished Human Text} and \textit{Humanized LLM Text} classes}.
Inspired by the conceptual framework from~\citet{bao2025-hart} and prior four-class fine-grained detection efforts such as DetectAIve~\citep{abassy-etal-2024-llm}, we analyze the four classes through the dual lenses of \textit{creator} and \textit{editor} and propose to enhance the modeling of creators' contributions for fine-grained detection.
As illustrated in \Fref{fig:research-scope}, the creator establishes the basic elements and logical flow, while the editor controls the linguistic expression and surface-level style of these elements.
For the pure human/LLM classes, differentiating the two roles is unnecessary; thus, conventional binary classifiers only need to obtain unified features to model human-LLM differences.
In contrast, the creator-editor collaboration modes for the two mixed classes are quite different: LLM-Polished Human Text originates from a human creator's framework and is subsequently refined by an LLM's stylistic surface, whereas Humanized LLM Text has an LLM-generated foundation but is then edited by humans to perturb LLM traits.
These divergent modes produce unique traits that are hard for unified features to capture, making it essential to look beyond the final actor and model the contributions of the creator and editor roles separately.

To address the four-class detection challenge, we propose the \textbf{R}hetorical \textbf{A}nalysis for \textbf{C}reator-\textbf{E}ditor Modeling (\textbf{RACE}) that explicitly models the distinct contributions of the creator and the editor.
RACE is grounded in the argument that an editor's influence is primarily manifested in the linguistic expression, while the creator's identity is deeply rooted in the logical organization and argumentative progression of the content.
To model the editor's role, RACE first segments the text into Elementary Discourse Units (EDUs) and extracts their semantic representations, which reflect surface-level linguistic choices and refinements.
To model the creator's role, RACE utilizes Rhetorical Structure Theory (RST)~\citep{mann1988rhetorical} to construct an EDU-based logical relation graph that characterizes the foundational organization of the text, highlighting the human-LLM creation differences stemming from their fundamental knowledge formation mechanisms.
The graph is then processed by rhetoric-guided message passing to propagate information to capture complex rhetorical dependencies, which produces a root pooling representation for final prediction. 
Our contributions are as follows:
\begin{itemize}

\item \textbf{Task Specificity:} We introduce a task-specific detection framework for the four-class setting that explicitly identifies \textit{LLM-Polished Human Text} and \textit{Humanized LLM Text}.

\item \textbf{Method:} We propose RACE, a creator-editor modeling approach that leverages Rhetorical Structure Theory to capture the creator's logical organization and the editor's linguistic refinements for fine-grained detection.

\item \textbf{Performance:} Extensive experiments demonstrate the superiority of RACE under the four-class fine-grained setting with low false alarms.
\end{itemize}

\section{Preliminaries}

\subsection{Rhetorical Structure Theory}

\begin{wrapfigure}{r}{0.42\linewidth}
  \vspace{-6em}
  \centering
  \includegraphics[width=\linewidth]{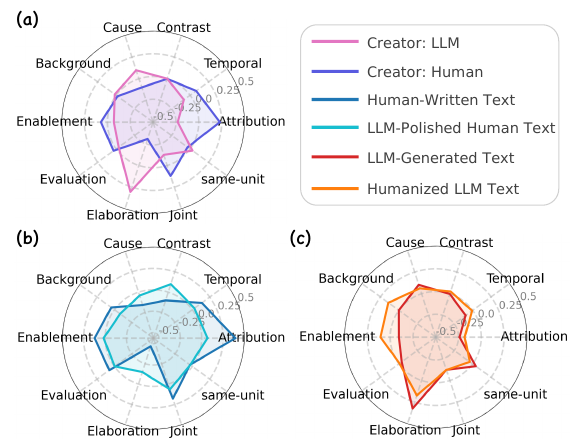}  
  \caption{Distribution of RST relations. (a) Divergence of Creators: Human creators build deeper rhetorical hierarchies (\eg, Attribution, Temporal), whereas LLMs produce flatter structures relying on surface-level relations (\eg, Elaboration, Cause). (b) LLM-Polished: underlying human architecture persists. (c) Humanized: underlying LLM architecture persists.}
  \label{fig:rst_radar}
  \vspace{-1em}
\end{wrapfigure}

Rhetorical Structure Theory (RST) is a descriptive framework for natural text organization, originally proposed to analyze how coherent discourse is constructed~\citep{mann1988rhetorical}.
RST models the hierarchical and functional dependencies between text spans, treating a text piece not as a linear sequence of words but as a structured tree of logical intentions.

The construction of a structured tree begins with segmenting the text into several spans called Elementary Discourse Units, which are typically clauses or phrases.
Then the text spans are linked through rhetorical relations (\eg, Elaboration, Contrast, Cause), resulting in an RST tree.
The RST tree can serve as a fingerprint of the creator's thought process.
Specifically, we posit that human and machine creators exhibit distinct structural signatures.
For humans, the writing process is inherently teleological, employing complex rhetorical relations such as clausal coordination to guide readers through a preset logical progression.
In contrast, LLMs, driven by auto-regressive probability, prioritize informational density over narrative logic, resulting in superficial structural signatures~\citep{reinhart2025llms}. 
By modeling logical organization, we can capture the intrinsic differences in how humans and machines architect their narratives.

\subsection{Motivating Analysis}

\begin{table}[t]
  \centering
  \small
  \begin{tabular}{llcc}
    \toprule
    \textbf{Reference} & \textbf{Target} & \textbf{Mean} & \textbf{Std} \\
    \midrule
    Human-Written & LLM-Polished & 0.92 & 0.08 \\
    Human-Written & LLM-Generated & 0.84 & 0.12 \\
    Human-Written & Humanized & 0.86 & 0.12 \\
    LLM-Generated & Human-Written & 0.84 & 0.12 \\
    LLM-Generated & LLM-Polished & 0.89 & 0.09 \\
    LLM-Generated & Humanized & 0.95 & 0.07 \\
    \bottomrule
  \end{tabular}
  \caption{Sample-level cosine similarity between rhetorical-relation frequency vectors on HART dataset.}
  \label{tab:creator_hypothesis_hart}
\end{table}

We conducted a preliminary statistical analysis on the distribution of RST relations across the HART dataset~\citep{bao2025-hart}.
Specifically, we adopt the Z-score to measure the deviation of each relation's frequency. For an RST relation $j$ in class $k$, the Z-score is calculated as $Z_{k,j} = (\bar{x}_{k,j} - \mu_j)/\sigma_j$, where $\bar{x}_{k,j}$ is the intra-class mean of relative frequency, and $\mu_j, \sigma_j$ are global mean and standard deviation. A value of $Z > 0$ indicates over-expression relative to the general population.

As visualized in \Fref{fig:rst_radar}~(a), human creators show a significant over-expression in \textit{Attribution} and \textit{Temporal}, indicating a discourse structure that is better grounded in external sources and more contextually organized.
LLM creators, conversely, exhibit strong spikes in \textit{Elaboration} and \textit{Cause}, suggesting a stronger reliance on local expansion, lacking the deep intertextual grounding found in human writing.
In \Fref{fig:rst_radar}~(b), even after LLMs' polishing, the text retains the high \textit{Attribution} features, which are more aligned with the human creator.
Similarly, \Fref{fig:rst_radar}~(c) shows that human editing fails to mask the underlying LLMs' structural signatures, as \textit{Elaboration} remain dominant.

Using each document's rhetorical-relation frequency vector (\eg, \{`Elaboration': 5, `Contrast': 2\} $\to [5,2]$), we compute pairwise cosine similarity between a source text and its variants to give a complementary sample-level analysis. \Tref{tab:creator_hypothesis_hart} indicates that texts produced by the same creator remain rhetorically closer to each other even after subsequent editing, further supporting the creator-editor hypothesis.

These findings indicate that the subsequent editing operation generally preserves the underlying logic of the creator, which shows the possibility of separately modeling the unique characteristics of humans and LLMs as creators or editors. In the next section, we will introduce rhetorical structure information to model the dual roles for fine-grained LLM-generated text detection.

\section{Proposed Method: RACE}
\label{sec:method}

\begin{figure*}[t]
  \includegraphics[width=\linewidth]{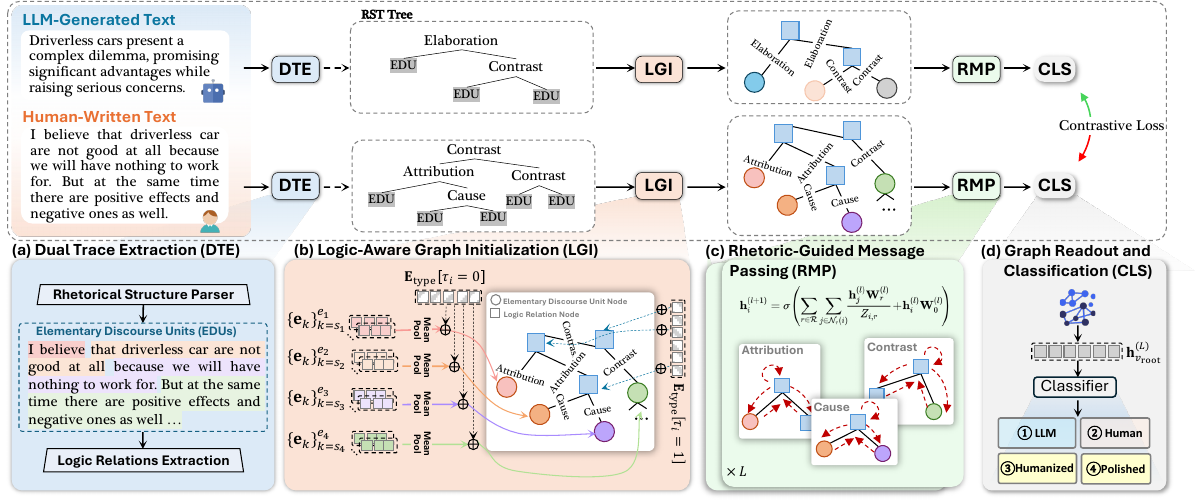}
  \caption{Overall architecture of \textbf{RACE}. Given a text piece, RACE \textbf{(a)} first captures both creator and editor traces through rhetorical structure construction and elementary discourse unit extraction. \textbf{(b)} These dual traces are then transformed into a logic-aware graph, where both linguistic expression and logical organization signals are encoded into node features via descendant span pooling and relation-aware projection. \textbf{(c)} Next, Rhetoric-Guided Message Passing propagates information through relation-specific aggregation with basis decomposition to capture complex rhetorical dependencies. \textbf{(d)} Finally, the global text representation is obtained via root pooling for classification.}  \label{fig:method}
\end{figure*}

To capture the dual trace of creator and editor for fine-grained LLM-generated text detection, we propose the logical-structure-aware detection framework, RACE.
As illustrated in \Fref{fig:method}, RACE consists of four key components: Dual Trace Extraction, Logic-Aware Graph Initialization
, Rhetoric-Guided Message Passing, and Graph Readout and Classification. Through these modules, RACE models the generative process through the dual lenses of linguistic expression and logical organization to improve fine-grained detection performance.

\subsection{Dual Trace Extraction}

To transform unstructured raw text into a structured logic-aware representation, we utilize the end-to-end RST parser developed by~\citet{chistova-2024-rst-parser}, which achieves superior performance in identifying hierarchical discourse dependencies.

Formally, the parsing process is defined as a mapping function $\mathcal{F}_{\mathrm{parse}}: D \rightarrow \mathcal{T}$.
Given an input text piece $D$, the parser outputs a binary constituency tree $\mathcal{T}$ that explicitly encodes the relation topology.
In this structure, the leaf nodes constitute the sequence of EDUs $\mathcal{V}_{\mathrm{edu}} = \{u_1, u_2, \dots, u_{|\mathcal{V}_{\mathrm{edu}}|}\}$, where each $u_i$ aligns with a specific continuous text span $[s_i, e_i]$.
Recursively, the internal nodes $\mathcal{V}_{\mathrm{rel}} = \{v_1, v_2, \dots, v_{|\mathcal{V}_{\mathrm{rel}}|}\}$ capture the logical organization by assigning a specific rhetorical label $r \in \mathcal{R}$ (\eg, Elaboration, Contrast) to the dependencies between sub-trees.
The tree $\mathcal{T}$ serves as the foundational skeleton, which is subsequently transformed into a logic-aware multi-relational graph to enable rhetoric-guided message passing.

\subsection{Logic-Aware Graph Initialization}
\label{sec:init}

Building upon the parsed tree $\mathcal{T}$, the text piece is formalized as a multi-relational graph $\mathcal{G} = (\mathcal{V}, \mathcal{E}, \mathcal{R})$, where $\mathcal{V} = \mathcal{V}_{\mathrm{edu}} \cup \mathcal{V}_{\mathrm{rel}}$. Each edge $e \in \mathcal{E}$ is represented as a triplet $(u, r, v)$, preserving the explicit dependency structure where relation nodes govern their constituent EDUs.

Furthermore, a hybrid strategy combining descendant span pooling with information bottleneck projection is proposed to initialize non-leaf nodes with semantically-informed representations, thus going beyond surface-level relation labels to encode richer contextual information.

\noindent\textbf{Descendant Span Pooling.}
For a text piece $D$ with tokens $\{t_1, \dots, t_K\}$, a pre-trained language model (PLM) is employed as the backbone to produce a sequence of contextualized embeddings $\mathbf{E} \in \mathbb{R}^{K \times d_{\mathrm{PLM}}}$.
The content representation $\mathbf{c}_i$ for any node $v_i \in \mathcal{V}$ is computed recursively:
\begin{equation}
  \label{eq:node_feature}
  \mathbf{c}_i = 
    \begin{cases} 
    \operatorname{MeanPool}(\{\mathbf{e}_k\}_{k=s_i}^{e_i}), & \text{if } v_i \in \mathcal{V}_{\mathrm{edu}} \\
    \frac{1}{|\mathcal{D}(v_i)|} \sum_{u \in \mathcal{D}(v_i)} \mathbf{c}_u, & \text{if } v_i \in \mathcal{V}_{\mathrm{rel}},
    \end{cases}
\end{equation}
where $\mathbf{e}_k$ is the $k$-th row of $\mathbf{E}$, and $\mathcal{D}(v_i) \subset \mathcal{V}_{\mathrm{edu}}$ denotes the set of all descendant nodes in the sub-tree rooted at $v_i$.
This strategy ensures that relation nodes are initialized with the global semantic centroid of the text segments they govern.

\noindent\textbf{Information Bottleneck Projection.}
Raw semantic embeddings often contain surface-level lexical noise irrelevant to structural authorship analysis.
To filter this redundancy, a dimension reduction strategy is adopted as an information bottleneck.
Specifically, the PLM embeddings are projected into a compact structural space of dimension $d_{\mathrm{feat}}$:
\begin{equation}
  \label{eq:node_project}
  \begin{gathered}
      \mathbf{c}'_i = \mathbf{c}_i + \mathbf{E}_{\mathrm{type}}[\tau_i],\\
      \mathbf{h}_i^{(0)} = \operatorname{Dropout} \left( \operatorname{LN} \left( \mathbf{c}'_i\mathbf{W}_{\mathrm{proj}}  + \mathbf{b}_{\mathrm{proj}} \right) \right),
  \end{gathered}
\end{equation}
where $\tau_i \in \{0, 1\}$ indicates the node type (EDU or Relation), $\mathbf{E}_{\mathrm{type}}$ is the learnable node type embedding table, $\mathbf{W}_{\mathrm{proj}} \in \mathbb{R}^{d_{\mathrm{PLM}}\times d_{\mathrm{feat}}}$ and $\mathbf{b}_{\mathrm{proj}}$ are the projection parameters, and $\operatorname{LN}$ signifies layer normalization.
This compression forces the model to distill only the most salient features required for the subsequent rhetoric-guided message passing. 

\subsection{Rhetoric-Guided Message Passing}

To learn the human-LLM differences over complex rhetorical dependencies, an $L$-layer Relational Graph Convolutional Network (RGCN; \citealp{schlichtkrull2018rgcn}) is adopted on the logic-aware graph.
Unlike vanilla GCNs that treat all edges uniformly~\citep{kipf2017gcn}, RGCN assigns relation-specific transformation matrices, allowing the model to learn distinct propagation rules for different rhetorical logics.

\noindent\textbf{Message Aggregation.}
In each layer $l$, the node representation is updated as:

\begin{equation}
  \label{eq:node_update}
  \!\mathbf{h}_i^{(l+1)}\!=\!\sigma\!\left(\!\sum_{r \in \mathcal{R}}\! \sum_{j \in \mathcal{N}_r(i)}\!\! \frac{\mathbf{h}_j^{(l)} \mathbf{W}_r^{(l)}}{Z_{i,r}} \!+\! \mathbf{h}_i^{(l)} \mathbf{W}_0^{(l)}\! \right)\!,\!\!\!\!
\end{equation}
where $\sigma(\cdot)$ denotes the activation function, $\mathbf{W}_r^{(l)}$ is the relation-specific weight matrix for relation $r$ in the $l$-th layer, $\mathcal{N}_r(i)$ is the set of neighbors under relation $r$, $Z_{i,r}$ is a normalization constant, and $\mathbf{W}_0^{(l)}$ handles the self-loop update.

\noindent\textbf{Basis Decomposition for Regularization.}
Given the large number of fine-grained rhetorical relations, learning $\mathbf{W}_r^{(l)}$ for each relation leads to parameter explosion and overfitting.
To constrain the weight space, basis decomposition is employed:
\begin{equation}
  \label{eq:basis_decomp}
  \mathbf{W}_r^{(l)} = \sum_{k=1}^{B} \alpha_{rk}^{(l)} \mathbf{V}_k^{(l)},
\end{equation}
where $\{\mathbf{V}_k^{(l)}\}_{k=1}^{B}$ is a set of shared basis matrices, and $\alpha_{rk}^{(l)}$ are learnable scalar coefficients unique to relation $r$.
This technique forces the model to learn the ``atomic'' components of rhetorical logic, improving generalization on sparse relations.

\subsection{Graph Readout and Classification}

As the logical structure is inherently hierarchical with a single root node $v_{\mathrm{root}}$ encompassing the entire text piece's rhetorical intent, a root pooling strategy is employed to capture the global text representation.
The global representation $\mathbf{z}_{\mathcal{G}}$ is directly extracted from the root node's final hidden state:
\begin{equation}
  \label{eq:readout}
  \mathbf{z}_{\mathcal{G}} = \mathbf{h}_{v_{\mathrm{root}}}^{(L)}.
\end{equation}

Finally, the global representation $\mathbf{z}_{\mathcal{G}}$ is passed to a classification head:
\begin{equation}
  \label{eq:clf}
  \begin{gathered}
      \tilde{\mathbf{z}} \!=\! \sigma \!\left( \operatorname{Dropout}(\mathbf{z}_{\mathcal{G}})\mathbf{W}_{\mathrm{in}} \!+\! \mathbf{b}_{\mathrm{in}} \right),\\
      \hat{y} \!=\! \operatorname{Softmax}\! \left( \operatorname{Dropout}(\tilde{\mathbf{z}})\mathbf{W}_{\mathrm{out}} \!+\! \mathbf{b}_{\mathrm{out}}\! \right),
  \end{gathered}
\end{equation}
where $\mathbf{W}_{\mathrm{in}}, \mathbf{W}_{\mathrm{out}}$ are weight matrices and $\mathbf{b}_{\mathrm{in}}, \mathbf{b}_{\mathrm{out}}$ are bias vectors, $\sigma$ is a non-linear activation function, and $\hat{y}$ is the predicted probability.

\noindent\textbf{Optimization.}
RACE is optimized using a joint loss that combines the supervised contrastive loss ~\citep{khosla2020cl} $\mathcal{L}_\mathrm{con}$ and the cross-entropy loss $\mathcal{L}_\mathrm{ce}$. The former is applied to the normalized feature representations, encouraging the model to learn a compact representation space. The joint loss function is $\mathcal{L}_\mathrm{total} = \mathcal{L}_\mathrm{con} + \mathcal{L}_\mathrm{ce}$.

\section{Experiments}
\label{sec:Experiments}

\subsection{Experimental Setup}

\noindent\textbf{Dataset.}
We use the HART~\citep{bao2025-hart} benchmark for evaluation due to its coverage of the desired categories.
However, the official release only contains validation and test partitions.
To enable supervised learning, we reorganized the data distribution (see \Apdxref{sec:dataset_details}) and performed a train/val/test split at the 70:20:10 ratio using stratified sampling across diverse domains (\eg, News, Writing, ArXiv, and Essay), which ensures distribution consistency across all partitions. We further report a stricter group-aware split, where all variants derived from the same base text are assigned to the same partition, in \Apdxref{sec:group_split}.

\noindent\textbf{Metrics.}
To evaluate the quality of the model's probability estimates independent of arbitrary decision thresholds, we prioritize metrics that assess the global ranking capability of the classifier rather than hard predictions.
Specifically, we adopt:
\begin{itemize}
    \item Macro-Averaged AUROC, which evaluates the probability that a randomly selected positive instance from any class is ranked higher than a randomly selected negative instance.
    \item TPR@1\% FPR (True Positive Rate at the 1\% False Positive Rate), which requires the detector to make precise judgments while avoiding false alarms~\cite{tpr-fpr}.
\end{itemize}

\noindent\textbf{Baselines.}
Since there is no task-specific method specifically designed for this 4-class setting, we establish baselines by adapting methods originally designed for two-/three-class settings.
For a reasonable comparison, we cover 12 learning-based or metric-based methods and tailor them to the fine-grained setting (More details in \Apdxref{sec:baseline_details}):

For learning-based methods, we include RoBERTa~\cite{openai-roberta-d}, RoBERTa-DANN~\citep{abassy-etal-2024-llm}, CoCo~\cite{liu-etal-2023-coco}, DeTeCtive~\citep{guo2024detective}, and LF-Motifs~\citep{kim-etal-2024-mfidf}. 
For all methods, we increase the number of entries in the classification head (\ie, the fully connected layer) from 2 to 4. 
CoCo and LF-Motifs incorporate discourse information, while DeTeCtive adopts contrastive learning, which is similar to our method in design.

For metric-based methods, we include Binoculars~\citep{hans2024binoculars}, Fast-DetectGPT (F-DetectGPT;~\citealp{fastdetectgpt}), and TDT~\citep{west2025tdt}, which typically produce scalars as predictions and thus are hard to directly extend to multi-class scenarios. Here, we conduct necessary modifications to obtain features via these methods by extracting the last-layer representation and appending a learnable MLP for four-class prediction.

\noindent\textbf{Implementation Details.}
For RACE, we use RoBERTa-base~\cite{roberta} as the backbone and only fine-tune the last layer while keeping the preceding layers frozen.
The extracted features are projected to a dimension of 128 to initialize node features. 
The graph component consists of an RGCN with $L=2$ layers, a hidden dimension of 512, and 10 bases for parameter regularization. The temperature $\tau$ is 0.07 for supervised contrastive loss.
We select the best validation checkpoint for testing. All experiments were conducted on a single NVIDIA RTX 4090 GPU.

\subsection{Main Results}

\begin{table*}[t]
    \small
    \centering
    \setlength{\tabcolsep}{1.5pt}
    \begin{tabular}{l c cccc@{\hspace{1.5em}} c@{\hspace{1.5em}}} 
        \toprule
        \multirow{2}{*}[-3pt]{\textbf{Method}} & \multirow{2}{*}[-3pt]{\textbf{AUROC}} & \multicolumn{5}{c}{\textbf{TPR@1\%FPR}} \\
        \cmidrule(lr){3-7}
         & & Human-Written & LLM-Polished & LLM-Generated & Humanized & \textbf{Avg} \\
        \midrule
        
        RoBERTa~\citep{openai-roberta-d} & 92.22 & \underline{99.36} & 68.06 & 63.14 & 70.92 & 75.37 \\
        RoBERTa$_\text{DANN}$~\citep{abassy-etal-2024-llm} & 96.17 & 96.88 & 75.03 & 48.89 & 71.78 & 73.14 \\
        CoCo~\citep{liu-etal-2023-coco} & 97.67 & \textbf{99.68} & \underline{75.77} & 63.93 & \textbf{79.43} & \underline{79.70} \\
        SeqXGPT~\citep{seqxgpt} & 89.87 & 98.38 & 15.23 & 14.32 & 31.68 & 39.90 \\
        DeTeCtive~\citep{guo2024detective} & 95.74 & 98.62 & 0.00 & 0.00 & \underline{77.23} & 43.96 \\
        LF-Motifs~\citep{kim-etal-2024-mfidf} & \textbf{98.20} & 96.68 & 69.61 & \underline{67.01} & 75.62 & 77.23 \\
        \midrule
        
        Binoculars$_\text{MLP}$~\citep{hans2024binoculars} & 79.15 & 29.49 & 7.34 & 4.37 & 5.50 & 11.70 \\
        Binoculars$_\text{C-T}$~\citep{bao2025-hart} & 50.03 & 0.00 & 0.00 & 0.00 & 0.00 & 0.00 \\
        F-DetectGPT~\citep{fastdetectgpt} & 61.70 & 0.00 & 3.37 & 26.27 & 0.09 & 7.70 \\
        F-DetectGPT$_\text{MLP}$~\citep{fastdetectgpt} & 73.69 & 3.12 & 3.87 & 29.35 & 3.96 & 10.8 \\
        F-DetectGPT$_\text{C-T}$~\citep{bao2025-hart} & 49.93 & 0.00 & 0.00 & 0.00 & 0.00 & 0.00 \\
        TDT$_\text{SVC}$~\citep{west2025tdt} & 57.16 & 2.88 & 2.37 & 3.58 & 0.50 & 2.33 \\
        \midrule
        
        \textbf{RACE (Ours)} & 
        \underline{\res{97.99}{0.13}} & 
        \res{99.04}{0.40} & 
        \textbf{\res{83.60}{1.61}} & 
        \textbf{\res{74.18}{0.95}} & 
        \res{75.41}{1.03} & 
        \textbf{\res{83.06}{0.57}} \\ 

        \bottomrule
    \end{tabular}
    \caption{
        Quantitative comparison of detection methods under the 4-class setting. For RACE, we report the results across three runs using different seeds in the format of the $\text{mean}_{\pm \text{std}}$. All values are reported in percentage (\%). \textbf{Bold} and \underline{underlined} values denote the best and second-best performance, respectively.
    }
    \label{tab:main_results}
\end{table*}

\Tref{tab:main_results} presents the quantitative comparison of RACE against baselines. We observe that:

\textbf{1) RACE achieves the highest average performance in TPR@1\%FPR.}
Specifically, RACE outperforms the best baseline CoCo by 3.36\% absolute with a low false alarm rate, indicating the effectiveness of the creator-editor dual modeling framework. 

\textbf{2) RACE outperforms closely related discourse-aware detection methods.} Similar to RACE, CoCo and LF-Motifs utilize discourse information: CoCo relies primarily on entity-coherence graphs to model inner- and inter-sentence relations; while LF-Motifs introduces statistical features of RST trees concatenated with Longformer embeddings. Though they outperform other compared baselines, CoCo struggles to capture the local stylistic shift when semantic entities remain unchanged, and LF-Motifs's statistical features are relatively shallow. In contrast, RACE leverages RGCN for message passing directly over the relational graph, thereby capturing the intrinsic structural topology and logical anomalies that shallow motifs fail to represent.

\textbf{3) Learning-based methods generally outperform metric-based ones for fine-grained classification.}
Metrics-based methods typically compress information into scalar values, which may be simple and effective for the binary setting, but the loss that such compression leads to also collapses the high-dimensional feature space necessary for the multi-class task. Aligned with the observation from~\citet{tpr-fpr}, we see poor performance for certain detectors, with TPR@1\%FPR as low as 0\%.
Even if we adopt several modifications to preserve more information, their performance still falls behind the learning-based methods, perhaps because the latter could entail the classification knowledge into well-trained parametric networks.

\subsection{Ablation Study}

\begin{table*}[t]
    \centering
    \small

    \begin{tabular}{l c cccc@{\hspace{2em}}c@{\hspace{2em}}}
        \toprule
        \multirow{2}{*}[-3pt]{\textbf{Method}} & \multirow{2}{*}[-3pt]{\textbf{AUROC}} & \multicolumn{5}{c}{\textbf{TPR@1\%FPR}} \\
        
        \cmidrule(l){3-7}
        
         & & Human-Written & LLM-Polished & LLM-Generated & Humanized & \textbf{Avg} \\
        \midrule
        
        \textbf{RACE} 
        & \res{97.99}{0.13}
        & \res{99.04}{0.40} & \res{83.60}{1.61} & \res{74.18}{0.95} & \res{75.41}{1.03} & \res{83.06}{0.57} \\
        
        \grayvar{w/o CL}
        & \res{97.73}{0.44} 
        & \res{98.21}{0.19} & \res{78.07}{1.06} & \res{69.10}{2.66} & \res{73.43}{1.59} & \res{79.70}{1.12} \\

        \grayvar{w/o Relation}
        & \res{96.78}{0.65} 
        & \res{97.42}{0.75} & \res{78.24}{2.08} & \res{65.35}{9.19} & \res{74.92}{1.74} & \res{78.98}{2.95} \\
        
        \grayvar{w/o RGCN}
        & \res{97.91}{0.20} 
        & \res{98.92}{0.19} & \res{82.27}{4.56} & \res{65.35}{6.71} & \res{74.42}{1.14} & \res{80.24}{2.07} \\
       \cmidrule(l){1-7}
        \grayvar{w/o Bottleneck}
        & \res{98.07}{0.22} 
        & \res{98.54}{0.26} & \res{80.82}{2.45} & \res{74.68}{0.86} & \res{75.74}{0.99} & \res{82.45}{0.60} \\
        \grayvar{w/o Basis}
        & \res{97.22}{0.74} 
        & \res{97.92}{0.26} & \res{83.39}{3.20} & \res{73.02}{5.46} & \res{74.58}{1.14} & \res{82.23}{1.74} \\
        
        \bottomrule
    \end{tabular}
    \caption{Ablation results ($\text{mean}_{\pm \text{std}}$) of RACE. The \textit{Bottleneck} represents the Information Bottleneck Projection in \eref{eq:node_project}, the \textit{Basis} represents the Basis Decomposition in \eref{eq:basis_decomp}, and \textit{w/o Relation} means removing the relation types on edges and adopting vanilla GCN~\citep{kipf2017gcn}.}
    \label{tab:ablation}
\end{table*}

As presented in~\Tref{tab:ablation}, the TPR@1\%FPR shows a clear drop when removing the involved components, confirming their individual benefits for improving fine-grained detection performance.
We notice the largest performance drop arises at the LLM-generated class, followed by the LLM-Polished class.
This is aligned with our intuition: The LLM-generated and polished samples are created by LLMs and humans, respectively, but share the same editor (\ie the LLM).
The degradation of \textit{w/o Relation} variant indicates that, without logical relations, vanilla GCN fails to capture patterns defining the role of creator and confirms the core advantage of RACE.
Without the contrastive learning that enhances the feature differences and the RGCN that explicitly models the dual roles of creator and editor, the detector might mix the characteristics of the two editor-similar types of samples. 

Furthermore, the results of \textit{w/o Basis} and \textit{w/o Bottleneck} validate the necessity of parameter efficiency and feature compression.
Specifically, removing the Basis Decomposition leads to a noticeable increase in performance variance (\ie, high standard deviation).
This means that the basis decomposition is an important regularizer because it forces weight sharing between similar relations. This stops over-parameterization and makes sure that optimization stays stable.
Meanwhile, the removal of the Information Bottleneck Projection causes a specific performance degradation on the LLM-Polished class. 
This corroborates that this module effectively prevents the model from overfitting to superficial patterns shared with the LLM editor and forces it to focus on the invariant features indicative of the human creator.

\begin{table}[t]
  \centering
  \small
  \begin{tabular}{lcc}
    \toprule
    \textbf{Variant} & \textbf{AUROC} & \textbf{Avg. TPR@1\%FPR} \\
    \midrule
    RACE (DMRST) & 97.64 & 82.93 \\
    RACE (IsaNLP) & 97.99 & 83.06 \\
    \bottomrule
  \end{tabular}
  \caption{Parser-robustness analysis.}
  \label{tab:parser_robustness}
  \vspace{-1em}
\end{table}

We also experiment with different RST parsers, IsaNLP~\citep{chistova-2024-rst-parser} and DMRST~\citep{liu-etal-2021-dmrst}, while keeping the downstream architecture fixed. \Tref{tab:parser_robustness} shows that RACE remains stable under parser replacement, suggesting that the core improvement comes from creator-editor modeling rather than artifacts of one parser.

\subsection{Further Analysis}

To provide a deeper insight into the effectiveness of our proposed method, we conduct a comprehensive comparison against CoCo and LF-Motifs, which are top-performing in the main experiments.

\noindent\textbf{Discriminability of Feature Representations.}
To quantitatively evaluate the discriminability of the learned representations, we employ two standard clustering validity indices: the Davies-Bouldin Index~(DBI)~\citep{dbi} and the Calinski-Harabasz Index~(CH)~\citep{ch}.
As shown in \Tref{tab:separation_metrics}, RACE achieves a lower DBI of $0.8042$ than CoCo ($0.9286$), indicating a better ratio of intra-cluster scatter to inter-cluster separation.
Regarding the CH index, RACE's score is nearly double that of CoCo, implying that explicitly modeling the creator/editor logic leads to a significantly more compact and distinct embedding space. The indices show the superiority of RACE in learning a discriminative feature space for fine-grained detection.

\begin{wrapfigure}{r}{0.47\linewidth}
  \centering
  \includegraphics[width=\linewidth]{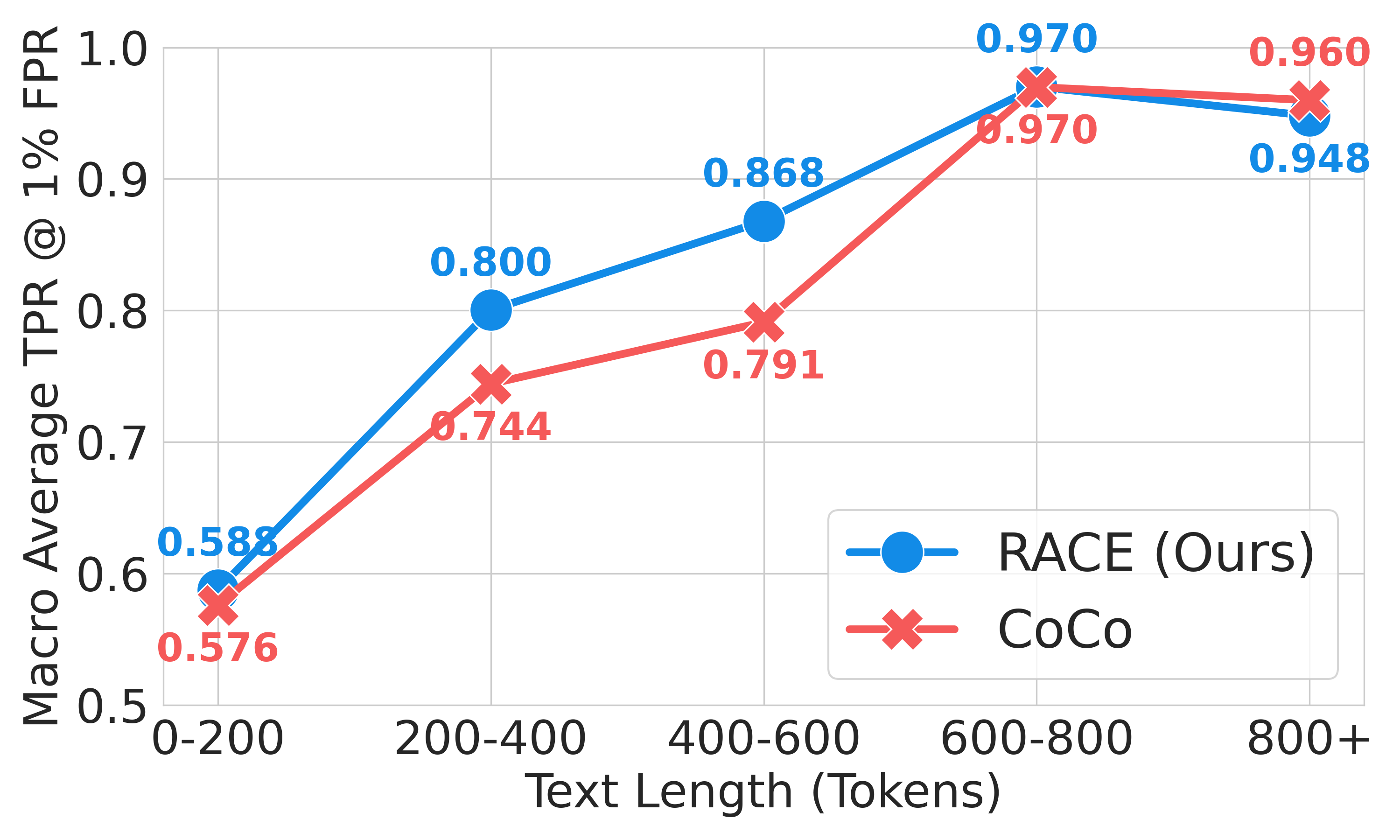}
  \caption{Analysis of detection performance of CoCo and our proposed RACE across varying text lengths.}
  \label{fig:metrics_by_len}
  \vspace{-1em}
\end{wrapfigure}

\noindent\textbf{Impact of Text Length Variations.}
We investigate how input text length affects detection performance.
\Fref{fig:metrics_by_len} illustrates the TPR@1\% FPR across different token length intervals.
While both methods perform well on texts longer than 600 tokens, RACE performs better for shorter texts with 200-600 tokens.
This advantage suggests that our graph-based approach is more efficient in capturing the nuanced differences introduced by creators and editors, with relatively limited information.

\noindent\textbf{Out-of-Distribution (OOD) Testing.}

\begin{table}[tbp]
    \centering
    \small
    \begin{tabular}{l c c}
        \toprule
        \textbf{Metric} & \textbf{CoCo} & \textbf{RACE} \\
        \midrule
        Davies-Bouldin Index ($\downarrow$) & 0.9286 & 0.8042 \\
        Calinski-Harabasz Index ($\uparrow$) & 2289.40 & 4333.32 \\
        \bottomrule
    \end{tabular}
    \caption{Quantitative evaluation of feature discriminability using clustering validity indices.}
    \label{tab:separation_metrics}
\end{table}

To assess the robustness of our method across different text genres, we extend the evaluation to a Leave-One-Domain-Out setting on the four domains in HART, including \textit{Arxiv}, \textit{Essay}, \textit{News}, and \textit{Writing}.
Excluding the preserved domain, the samples in the remaining three domains are split into the training and validation sets with a ratio of 9:1.
From \Tref{tab:ood}, we see that RACE outperforms CoCo and LF-Motifs in most cross-domain scenarios, particularly for structured genres like research articles and essays.
For CoCo, the entity distributions are highly domain-dependent, and when transferred to an unseen domain, the learned entity patterns fail to reflect the new inductive bias, leading to performance degradation.
Differently, RACE relies on both linguistic expression and logical organization and forms a more comprehensive view, thus enhancing the OOD generalizability.

\begin{table}[tbp]
  \centering
  \small
  \setlength{\tabcolsep}{5pt} 
    \begin{tabular}{clcc}
    \toprule
    \textbf{Domain} & \textbf{Method} & \textbf{AUROC} & \textbf{Avg. TPR@1\%FPR} \\
    \midrule
    \multirow{3}{*}{Arxiv} 
      & CoCo & 93.67 & 55.91 \\
      & LF-Motifs & \underline{94.07} & \underline{58.64} \\
      & RACE & \textbf{96.61} & \textbf{76.28} \\
    \midrule
    \multirow{3}{*}{Essay} 
      & CoCo & 89.84 & 28.72 \\
      & LF-Motifs & \underline{91.12} & \underline{47.85} \\
      & RACE & \textbf{95.88} & \textbf{59.73} \\
    \midrule
    \multirow{3}{*}{News} 
      & CoCo & 89.35 & \underline{39.36} \\
      & LF-Motifs & \underline{91.04} & 35.36 \\
      & RACE & \textbf{92.69} & \textbf{44.30} \\
    \midrule
    \multirow{3}{*}{Writing} 
      & CoCo & 83.03 & \textbf{30.85} \\
      & LF-Motifs & \underline{84.73} & \underline{31.59}\\
      & RACE & \textbf{86.20} & 30.84 \\
    \bottomrule
    \end{tabular}
  \caption{Performance comparison under the OOD setting. We employ a Leave-One-Domain-Out protocol where the model is trained on three domains and tested on the fourth unseen domain (Column 1).}
  \label{tab:ood}
    \vspace{-1em}
\end{table}

\section{Related Work}
\label{sec:related_works}

\subsection{Binary Classification}

The binary classification is to judge whether a text piece is generated by the LLM.
Under this setting, a detector assumes that \textbf{1)} all samples are either purely written by humans or generated by LLMs; or \textbf{2)} any text involving LLMs belongs to the ``LLM'' class.
To model the differentiable signals, most existing methods focus on developing distribution-aware metrics like token probabilities~\citep{gltr}, token ranks~\citep{detectllm}, or their combinations~\citep{miralles2026not}, as these metrics reflect the disparities of human and LLM texts in word use.
To instantiate this, researchers utilized various signals to manifest or amplify such disparities.
For example, regeneration-based methods query an LLM with the given text to measure how similar the output is to the input, reflecting the familiarity the queried LLM is with the given text~\cite{zhu-etal-2023-beat,raidar,wrote}. The variants leverage multiple regenerations to calculate probability divergence~\citep{yang2024dnagpt} or consider the impact of the prompts~\citep{dpic}.
Perturbation-based methods operate in the embedding space, assuming that LLM-generated text resides in negative curvature regions of log-likelihood~\citep{mitchell2023detectgpt,fastdetectgpt}.
Another research line directly learns stylistic representations using supervised learning~\citep{openai-roberta-d,guo2023hc3,soto2024fewshot,guo2024detective}. Among the supervised methods, \citet{liu-etal-2023-coco} and \citet{kim-etal-2024-mfidf} capture deeper linguistic structures and discourse coherence. To model the dual roles of creator and editor, we follow this line by introducing the RST tree, which deepened the understanding of discourse-level information.

\vspace{-1em}
\subsection{Fine-grained Classification}

Fine-grained classification is to differentiate the specific involvement of the human and the LLM to satisfy the regulation and forensics needs.
From an identity perspective, some works attribute the given text to a specific LLM, thereby formulating a model attribution task~\cite{sniffer,shi2024ten,li2025continual}.

Recently, human-LLM collaborative writing has become prevalent, and more works focus on differentiating the behavior and its extent in the resulting mixed text.
\citet{zeng2024towards} identify the transition points between human and LLM texts by comparing the prototypes of neighboring text snippets.
SeqXGPT~\citep{seqxgpt} treats detection as a sequence labeling problem for precise localization within mixed texts.
Other works set a third category to represent the mixed text.
FAIDSet~\citep{ta2025faid} categorizes text into three distinct classes: human-written, LLM-generated, and collaborative, with specific labels for LLM-polishing and LLM-continuation. APT-Eval~\citep{saha2025almost} considers the different levels of LLM polishing.
\citet{zhou-etal-2024-humanizing} study the adversarial behavior named humanizing, typically to bypass LLM text detectors to earn unethical advantages.
At the document level, \citet{zhang-etal-2024-mixset} constructs MixSet, pioneering document-level differentiation between machine-revised human text and human-revised machine text through diverse editing operations such as Polish and Humanize.
\citet{abassy-etal-2024-llm} fine-tune pretrained language models for a four-class authorship setting and release the DetectAIve demo system.
Recently, \citet{bao2025-hart} designed a detector that explicitly decouples text into content and expression dimensions, identifying LLM artifacts primarily in the expression layer.

However, they mainly focus on the linguistic expression, which represents the synthesized outcome after human-LLM collaboration, thus failing to reveal role-specific traits. In contrast, our proposed RACE incorporates both linguistic expression and the logical organization signals through rhetoric-guided graph learning, which models the generative process through the dual lenses of creator and editor, enabling superior performance on fine-grained detection.

\section{Conclusion}
\label{sec:Conclusion}

We studied the four-class setting in fine-grained LLM-generated text detection, to distinguish human-written text, LLM-generated text, LLM-polished human text, and humanized LLM text. 
We modeled the dual roles of creator and editor through rhetorical structure construction and elementary discourse unit extraction, and designed the detector, RACE.
By building the logic-aware graph and performing rhetoric-guided message passage, RACE outperformed 12 baselines on the HART benchmark with a low false alarm rate.

\section*{Limitations}
In this paper, we conducted an initial exploration to perform the complex four-class task for fine-grained LLM-generated text detection. Despite the effectiveness of the proposed method RACE, we identify the following limitations:

1) We only conduct experiments on one public benchmark (\ie, HART) because it is the only accessible dataset suitable for the four-class setting when we conducted this study. The performance of RACE on other languages, domains, and genres that HART does not cover remains unknown.

2) There is still room for RACE in terms of absolute performance improvement to satisfy the requirements for commercial use. Therefore, it is not recommended to directly take subsequent actions according to RACE's predictions without additional manual checks. Further research in this direction is advocated.

3) Though the four-class setting has been complex, there indeed exists the possibility that a text piece is the result of a longer editing sequence. A recent study began to consider sample editing multiple times by different LLMs~\cite{detree}, but we focus more on constructing the basic setting and thus did not explore this kind of effect.

\bibliography{main}

\appendix

\section{More Details of Experiment Setups}

\subsection{Dataset Details}
\label{sec:dataset_details}

We reconstructed the dataset by processing the raw JSON files from the HART benchmark, merging the original development and test splits into a unified corpus. To support our classification framework, we implemented a parsing pipeline that assigns fine-grained labels based on the unique record identifiers (\code{id}) and metadata fields (\code{content\_source}, \code{language\_source}) of each entry.
The mapping logic is defined as follows:
\begin{itemize}
    \item Human-Written Text: Identified by base record IDs lacking derivative prefixes (\eg, \code{gen/}, \code{rep/}), representing the original, unaltered human authorship.
    \item LLM-Generated Text: Primarily derived from records prefixed with \code{gen/}, where the \code{content\_source} indicates a machine origin (\eg, \code{machine:gpt-4}). We also map LLM-to-LLM revision chains (prefixed \code{hum/gen/} where the reviser is another model) to this category, treating them as fully synthetic content.
    \item LLM-Polished Human Text: Extracted from records prefixed with \code{rep/}, where the \code{language\_source} (tagged as \code{rephrase:}) indicates that an original human text was refined by an LLM.
    \item Humanized LLM Text: Identified from records prefixed with \code{hum/gen/}, specifically filtering for instances where the \code{language\_source} is tagged as \code{humanize:human} or \code{humanize:tool}. This captures the distinct scenario of synthetic text subsequently edited by human annotators or grammar correction tools.
\end{itemize}

No identical text appears across the resulting train/val/test partitions because each record is unique after reconstruction. To complement the default variant-level random split, we further report a stricter group-aware split in \Apdxref{sec:group_split}.

\begin{table}[tbp]
    \centering
    \small
    \setlength{\tabcolsep}{3pt}
    \begin{tabular}{l l r r r r}
        \toprule
        \textbf{Domain} & \textbf{Category} & \textbf{Train} & \textbf{Val} & \textbf{Test} & \textbf{Total} \\
        \midrule
        
        \multirow{4}{*}{\textbf{Arxiv}} 
        & Human-Written & 700 & 100 & 200 & 1,000 \\
        & LLM-Polished   & 700 & 100 & 200 & 1,000 \\
        & LLM-Generated & 1,229 & 174 & 352 & 1,755 \\
        & Humanized    & 172 & 25 & 48 & 245 \\
        \cmidrule{1-6}
        
        \multirow{4}{*}{\textbf{Essay}} 
        & Human-Written & 700 & 100 & 200 & 1,000 \\
        & LLM-Polished     & 700 & 100 & 200 & 1,000 \\
        & LLM-Generated & 1,220 & 175 & 349 & 1,744 \\
        & Humanized    & 179 & 25 & 52 & 256 \\
        \cmidrule{1-6}

        \multirow{4}{*}{\textbf{News}} 
        & Human-Written & 700 & 100 & 200 & 1,000 \\
        & LLM-Polished     & 700 & 100 & 200 & 1,000 \\
        & LLM-Generated & 1,229 & 175 & 354 & 1,758 \\
        & Humanized    & 169 & 25 & 48 & 242 \\
        \cmidrule{1-6}
        
        \multirow{4}{*}{\textbf{Writing}} 
        & Human-Written & 700 & 100 & 200 & 1,000 \\
        & LLM-Polished     & 700 & 99 & 201 & 1,000 \\
        & LLM-Generated & 1,211 & 175 & 342 & 1,728 \\
        & Humanized    & 191 & 27 & 54 & 272 \\
        \midrule
        
        \multicolumn{2}{l}{\textbf{Total}} & \textbf{11,200} & \textbf{1,600} & \textbf{3,200} & \textbf{16,000} \\
        \bottomrule
    \end{tabular}
    \caption{Statistics of the resplit HART dataset.}
    \label{tab:dataset_stats}
\end{table}

\Tref{tab:dataset_stats} presents the detailed statistics of the dataset.
The LLMs used in the data include Claude-3.5-Sonnet~\citep{claude-3p5-sonnet}, GPT-3.5-Turbo, GPT-4o~\citep{openai-gpt4o}, Gemini-1.5-Pro~\citep{team2024gemini1p5}, Llama-3.3-70b-Instruct~\citep{grattafiori2024llama-3p3}, and Qwen-2.5-72b-Instruct~\citep{qwen2025qwen25technicalreport}.

\subsection{Metrics Calculation}
\label{sec:metric_details}

We adopt macro AUROC and TPR at 1\% FPR for the main experiments. Let $\mathcal{C} = \{1, \dots, C\}$ be the set of classes (here, $C=4$).
For each class $c \in \mathcal{C}$, let $y_{i,c} \in \{0, 1\}$ denote the binary label and $\hat{p}_{i,c} \in [0, 1]$ the predicted probability for the $i$-th sample.
We treat the multi-class problem as $C$ independent binary classification tasks (One-vs-Rest).
The Macro-AUROC is defined as:
\begin{equation}
  \label{eq:auroc}
  \text{Macro-AUROC} = \frac{1}{C} \sum_{c=1}^{C} \text{AUROC}(y_{\cdot, c}, \hat{p}_{\cdot, c}),
\end{equation}
where $\text{AUROC}(\cdot, \cdot)$ denotes the standard area under the receiver operating characteristic curve for each binary target.
The macro-averaged TPR at the 1\% FPR is defined as:
\begin{equation}
  \label{eq:tpr_at_1_fpr}
  \text{TPR}@1\%\text{FPR} = \frac{1}{C} \sum_{c=1}^{C} \text{TPR}_c(\tau_c),
\end{equation}
subject to:
\begin{equation}
  \tau_c = \min \{ \tau \in [0, 1] \mid \text{FPR}_c(\tau) \le 0.01 \}.
\end{equation}
Here, $\text{TPR}_c(\tau)$ and $\text{FPR}_c(\tau)$ represent the true positive and false positive rates for class $c$ at threshold $\tau$, respectively.
We use macro averaging to highlight the influence of the minority class in the dataset.

\subsection{More Implementation Details}
\label{sec:baseline_details}

\subsubsection{RACE}

\paragraph{Dual Trace Extraction}
We employ the IsaNLP RST Parser\footnote{\url{https://github.com/tchewik/isanlp_rst}} proposed by \citet{chistova-2024-rst-parser}, which maps relations to a unified set of 18 coarse-grained classes.
The released checkpoint we used can be found in their HuggingFace repository\footnote{\url{https://huggingface.co/tchewik/isanlp_rst_v3/tree/rstdt}}.
The relation types considered in our study include Attribution, Background, Cause, Comparison, Condition, Contrast, Elaboration, Enablement, Evaluation, Explanation, Joint, Manner-Means, Same-unit, Summary, Temporal, Textual-organization, Topic-Change, and Topic-Comment.

\paragraph{Backbone Model}
The pretrained RoBERTa-base model can be downloaded from Facebook AI's HuggingFace page\footnote{\url{https://huggingface.co/FacebookAI/roberta-base}}.

\paragraph{Optimization}
Let $\mathcal{B} = \{(x_i, y_i)\}_{i=1}^{N}$ denote a mini-batch of $N$ input samples, where $x_i$ represents the input text and $y_i \in \{1, \dots, C\}$ is the corresponding ground-truth label, with $C=4$ representing the number of classes.
The Supervised Contrastive Loss $\mathcal{L}_\mathrm{con}$ is formulated as:
\begin{equation}
  \label{eq:loss_con}
  \!\!\mathcal{L}_\mathrm{con}\!=\!\!\sum_{i \in I}\! \frac{-1}{|P(i)|} \!\!\sum_{p \in P(i)}\!\! \log\! \frac{e^{\mathbf{z}_{\mathcal{G}}^i \cdot \mathbf{z}_{\mathcal{G}}^p/ \tau}}{\sum_{a \in A(i)} e^{{\mathbf{z}_{\mathcal{G}}^i \cdot \mathbf{z}_{\mathcal{G}}^a} / \tau}}.
\end{equation}
Here, $\mathbf{z}_{\mathcal{G}}^i$ is the feature vector extracted by \eref{eq:readout} for the $i$-th sample, $I \equiv \{1, \dots, N\}$ is the set of indices in the batch. $A(i) \equiv I \setminus \{i\}$ represents the set of all indices excluding the anchor $i$. The set $P(i) \equiv \{p \in A(i) : y_p = y_i\}$ denotes the set of indices for positive samples sharing the same class label as $i$, and $|P(i)|$ is its cardinality. The symbol $\tau \in \mathbb{R}^+$ is a temperature parameter that controls the smoothness of the distribution.
For the Cross-Entropy Loss $\mathcal{L}_\mathrm{ce}$, we apply it on the classifier output $\hat{y}_i$ calculated by \eref{eq:clf}, formulated as:
\begin{equation}
  \label{eq:loss_ce}
  \mathcal{L}_\mathrm{ce} = - \frac{1}{N} \sum_{i=1}^{N} \sum_{c=1}^{C} \mathds{1}_{[y_i = c]} \log(\hat{y}_{i,c}),
\end{equation}
where $\mathds{1}_{[\cdot]}$ is the indicator function, $y_i$ is the ground-truth label, and $\hat{y}_{i,c} \in [0,1]$ is the predicted probability for class $c$.

\subsubsection{Baselines}

Given the absence of task-specific architectures specifically designed for this four-class classification framework, we selected representative methods from both learning-based and metric-based paradigms and adapted them to our reorganized HART dataset.

\paragraph{Learning-based Methods Adaptation.}
We modified the output dimensions of the classification heads to support four categories while retaining the original model architectures. Specifically:
\begin{itemize}
    \item \textbf{RoBERTa and RoBERTa-DANN:} We fine-tuned RoBERTa with a 4-way classifier. For RoBERTa-DANN, we additionally attached a domain-adversarial objective following DetectAIve~\citep{abassy-etal-2024-llm} while keeping the same 4-way prediction head.
    \item \textbf{CoCo and LF-Motifs:} We adjusted the final classification layer to output four class probabilities and optimize the learning rates on the training set to ensure convergence while keeping other hyperparameters consistent with the original implementations. Specifically, for LF-Motifs, we re-extracted the single, double, and triple triads from the HART corpus to reconstruct the features required for the four-class scenario.
    \item \textbf{SeqXGPT:} We reformulated the training objective by removing the Conditional Random Field (CRF) layer and discarding the fine-grained sequence labeling prefixes (\ie, B-, M-, E-, S-). Instead, the model was trained to predict one of the four category labels for each token directly. During inference, we maintained the original majority voting mechanism, aggregating token-level predictions to determine the sentence-level label.
    \item \textbf{DeTeCtive:} We extended the contrastive learning objective by expanding the sample definitions in the contrastive loss from a binary distinction (LLM v.s. Human) to the target four distinct categories. All other training configurations followed the original paper.
\end{itemize}

\paragraph{Metric-based Methods Adaptation.}
Since metric-based detectors are originally designed for binary classification via thresholding, we adapted them by leveraging their intermediate signals:
\begin{itemize}
    \item \textbf{Multi-interval Thresholding (F-DetectGPT):} We adapted Fast-DetectGPT by discretizing its probabilistic curvature score into four intervals using three empirical thresholds ($0.5$, $0.8$, and $1.2$). These intervals correspond to Human, LLM-Polished, Humanized, and LLM-Generated, respectively.
    \item \textbf{Feature Fusion with MLP (Binoculars$_\text{MLP}$ and F-DetectGPT$_\text{MLP}$):} We introduced an MLP classifier for each method independently. For Binoculars, we concatenated $\log \text{PPL}$ and $\log \text{X-PPL}$ to form the input feature vector; for Fast-DetectGPT, we used the curvature score as a single-dimensional feature. Both were then fed into their respective MLPs for four-class prediction.
    \item \textbf{Decoupled Content-Expression Judgment (Binoculars$_\text{C-T}$ and F-DetectGPT$_\text{C-T}$):} Following HART~\citep{bao2025-hart}, we employed a decoupled detection strategy. We performed binary classification independently on the content and expression dimensions. The final label is derived from the combination of these two binary outcomes: Human (Content: Human, Expression: Human), LLM-Generated (Content: LLM, Expression: LLM), LLM-Polished (Content: Human, Expression: LLM), and Humanized (Content: LLM, Expression: Human). For the expression dimension, we used the text to be tested following the best-performing setting ($C_2$-$T$) reported in HART; for the content dimension, we utilized the \code{content} provided in the original HART dataset.
\end{itemize}

\section{Additional Experimental Results}
\label{sec:more_rets}

\subsection{Supplementary Quantitative Comparison}

\begin{table}[t]
  \centering
  \small
  \begin{tabular}{@{\hspace{2em}}lc}
    \toprule
    \textbf{Method} & \textbf{Avg. TPR@5\%FPR} \\
    \midrule
    RoBERTa       & 92.53\\
    CoCo          & \underline{94.13}\\
    SeqXGPT       & 54.41\\
    DeTeCtive     & 92.82\\
    LF-Motifs     & 91.90\\
    \midrule
    Binoculars$_\text{MLP}$ & 28.00\\
    Binoculars$_\text{C-T}$ & 0.34\\
    F-DetectGPT & 13.99\\
    F-DetectGPT$_\text{MLP}$ & 23.27\\
    F-DetectGPT$_\text{C-T}$ & 0.34\\
    TDT$_\text{SVC}$ & 16.37 \\
    \midrule
    RACE (Ours)   & \textbf{94.41}\\
    \bottomrule
  \end{tabular}%
  \caption{Performance comparison of different detection methods evaluated using Avg. TPR@5\%FPR.}
  \label{tab:tpr_at_5_fpr}
\end{table}

\Tref{tab:tpr_at_5_fpr} reports the performance under a more relaxed constraint of 5\% FPR.
Our method consistently maintains the leading position with an average TPR of $94.41\%$.
This consistent superiority across different thresholds highlights the strong discriminative power of our model, which ensures a high safety margin and demonstrates its reliability for high-precision applications.

\subsection{Stricter Group-Aware Split}
\label{sec:group_split}

Beyond the default random split over variants, we additionally evaluate a stricter split that groups all variants of one base text into the same partition while preserving domain boundaries. The results are shown in \Tref{tab:group_split}.

We see that the core finding of the main experiment still holds true under this data split.
Even when different models discuss similar topics, the organizational logic of the content differs. This distinction is precisely the information our creator-editor modeling aims to capture, making our proposed RACE unlikely to be influenced by topic information.

\begin{table*}[t]
  \centering
  \small
  \setlength{\tabcolsep}{3pt}
  \begin{tabular}{l c cccc@{\hspace{2em}} c@{\hspace{2em}}} 
    \toprule
    \multirow{2}{*}[-3pt]{\textbf{Method}} & \multirow{2}{*}[-3pt]{\textbf{AUROC}} & \multicolumn{5}{c}{\textbf{TPR@1\%FPR}} \\
    \cmidrule(lr){3-7}
     & & Human-Written & LLM-Polished & LLM-Generated & Humanized & \textbf{Avg} \\
    \midrule
    RoBERTa & \textbf{98.39} & \textbf{99.72} & 72.63 & 70.18 & \textbf{71.69} & 78.55 \\
    CoCo & 97.51 & 99.33 & 77.48 & 44.03 & 65.00 & 71.46 \\
    LF-Motifs & 97.13 & 94.95 & 39.19 & 55.74 & 61.45 & 62.83 \\
    RACE (Ours) & 96.59 & 99.13 & \textbf{86.25} & \textbf{75.20} & 67.00 & \textbf{81.90} \\
    \bottomrule
  \end{tabular}
  \caption{Results under a stricter group-aware split that keeps all variants of the same base text in one partition.}
  \label{tab:group_split}
\end{table*}




\subsection{Efficiency Analysis}

\begin{table}[t]
  \centering
  \small
  \begin{tabular}{lccc}
    \toprule
    \textbf{Method} & \textbf{Training} & \textbf{Inference} & \textbf{Params} \\
    \midrule
    RoBERTa       & 2.119 & 220.8 & 125.2 \\
    CoCo          & 2.138 & 32.4  & 125.6 \\
    LF-Motifs     & 0.587 & 50.6  & 148.8 \\
    RACE (Ours)   & 1.071 & 90.0  & 128.6 \\
    \bottomrule
  \end{tabular}%
  \caption{Comparison of efficiency and model size. Training time (hours, \textbf{h}), inference throughput (\textbf{samples/s}), and model size (million parameters, \textbf{M}) are reported. Data preprocessing is excluded from both training time and inference throughput.}
  \label{tab:efficiency}
\end{table}

\Tref{tab:efficiency} presents the training time, inference throughput, and the number of parameters for the four top-performing methods.
While LF-Motifs appears to achieve the lowest training time, primarily due to its utilization of the optimized Hugging Face Trainer\footnote{\url{https://huggingface.co/docs/transformers/en/trainer}}, it requires a heavy data preprocessing phase to extract single, double, and triple triads. This extraction process takes over three hours, significantly longer than other compared methods.
Our proposed RACE consumes relatively low training and inference time with a comparable scale of model parameters, confirming its efficiency for model preparation and deployment in reality.

\section{Reproducibility}

The code is available at the following GitHub repository for reproducibility needs: \url{https://github.com/ICTMCG/RACE}.

\section{Future Work}

While existing methods have struggled to address the nuanced regulatory requirements of specific domains like academic writing, our work RACE demonstrates that decoupling the roles of creator and editor is a promising direction for next-generation LLM-generated text detection.
We foresee three pathways for future exploration motivated by this dual-role paradigm:
\begin{itemize}
    \item \textbf{Logic-Based Model Attribution:} Current attribution methods often rely on surface-level token probability distributions, which are fragile to simple editing.
    Future research could adapt RACE’s graph-based topological features to fingerprint the unique logical thought processes of specific LLMs, potentially enabling attribution even after heavy human polishing.
    \item \textbf{Fine-Grained EDU-Level Detection:} Future works could introduce EDU-level annotations or weakly supervised objectives to identify exact EDUs where a human editor intervenes in LLM-generated drafts, providing granular evidence for academic integrity investigations.
    \item \textbf{Adversarial Logic Defense:} As LLMs become capable of mimicking human rhetorical structures, the ``arms race'' will shift from lexical to logical obfuscation.
    Future logic-aware adversarial attacks where prompts explicitly request structural restructuring and corresponding defense mechanisms can be explored.
\end{itemize}

\end{document}